\documentclass[10pt,twocolumn]{article}

\usepackage{cvpr}
\usepackage{times}
\usepackage{epsfig}
\usepackage{graphicx}
\usepackage{amsmath}
\usepackage{amssymb}
\usepackage[inline]{enumitem}
\usepackage{adjustbox}    % adjustbox
\usepackage{booktabs}    % advanced hlines
\usepackage{multirow}        % multirow for table
\usepackage{color,colortbl}

\usepackage[linesnumbered,ruled]{algorithm2e} % algorithm

 \usepackage[pagebackref=true,breaklinks=true,letterpaper=true,colorlinks,bookmarks=false]{hyperref}
\hypersetup{
	colorlinks=true,
	linkcolor=magenta,
	filecolor=green,      
	urlcolor=magenta,
	citecolor=cyan,
}

\newcommand{\argmin}{\operatornamewithlimits{argmin}}
\newcommand{\sth}{^{\text{th}}}
\newcommand{\srcd}{\mathcal{D}^{s}}
\newcommand{\tgtd}{\mathcal{D}^{t}}

\newcommand{\model}{\mathcal{M}}
\newcommand{\region}{\mathbb{R}}

\newcommand{\srcx}{x^{s}}
\newcommand{\srcy}{y^{s}}
\newcommand{\nsrc}{N^s}
\newcommand{\ntgt}{N^t}
\newcommand{\tgtx}{x^{t}}
\newcommand{\tgty}{y^{t}}

\newcommand{\cL}{\mathcal{L}}

\newcommand{\amazon}{\texttt{A}}
\newcommand{\dslr}{\texttt{D}}
\newcommand{\webcam}{\texttt{W}}

 \cvprfinalcopy % *** Uncomment this line for the final submission
\def\cvprPaperID{6592} % *** Enter the CVPR Paper ID here

\ifcvprfinal\fi

\begin{document}
    \title{$d$-SNE: Domain Adaptation using Stochastic Neighborhood Embedding}
\author{Xiang Xu$^{\dagger}$ \thanks{Equal contribution. This work was done during Xiang's internship at AWS AI.},  
	Xiong Zhou $^{\ddagger}$\footnotemark[1], 
	Ragav Venkatesan$^{\ddagger}$, 
	Gurumurthy Swaminathan$^{\ddagger}$, 
	Orchid Majumder$^{\ddagger}$ \\
	$^{\dagger}$Computational Biomedicine Lab, University of Houston, Houston, USA\\
	$^{\ddagger}$  AWS AI, Seattle, USA\\
	\tt\small{$^{\dagger}$xxu18@central.uh.edu, $^{\ddagger}$\{xiongzho,ragavven,gurumurs,orchid\}@amazon.com}
}
\maketitle

	\begin{abstract}
	%On the one hand, deep neural networks are effective in learning large datasets. 
	%On the other, they are inefficient with their data usage. 
	Deep neural networks often require copious amount of labeled-data to train their scads of parameters. 
	Training larger and deeper networks is hard without appropriate regularization, particularly while using a small dataset. 
	Laterally, collecting well-annotated data is expensive, time-consuming and often infeasible. 
	A popular way to regularize these networks is to simply train the network with more data from an alternate representative dataset. 
	This can lead to adverse effects if the statistics of the representative dataset are dissimilar to our target.
	This predicament is due to the problem of domain shift. 
	Data from a shifted domain might not produce bespoke features when a feature extractor from the representative domain is used. 
	%Several techniques of domain adaptation have been proposed in the past to solve this problem. 
	In this paper, we propose a new technique  ($d$-SNE) of domain adaptation that cleverly uses stochastic neighborhood embedding techniques and a novel modified-Hausdorff distance. 
	The proposed technique is learnable end-to-end and is therefore, ideally suited to train neural networks. 
	Extensive experiments demonstrate that $d$-SNE outperforms the current states-of-the-art and is robust to the variances in different datasets, even in the one-shot and semi-supervised learning settings. 
	$d$-SNE also demonstrates the ability to generalize to multiple domains concurrently. 
\end{abstract}

%%%%%%%%% BODY TEXT
\section{Introduction}
\label{SEC:intro} 

\begin{figure}[!h]
	\centering
	\includegraphics[width=0.75\linewidth]{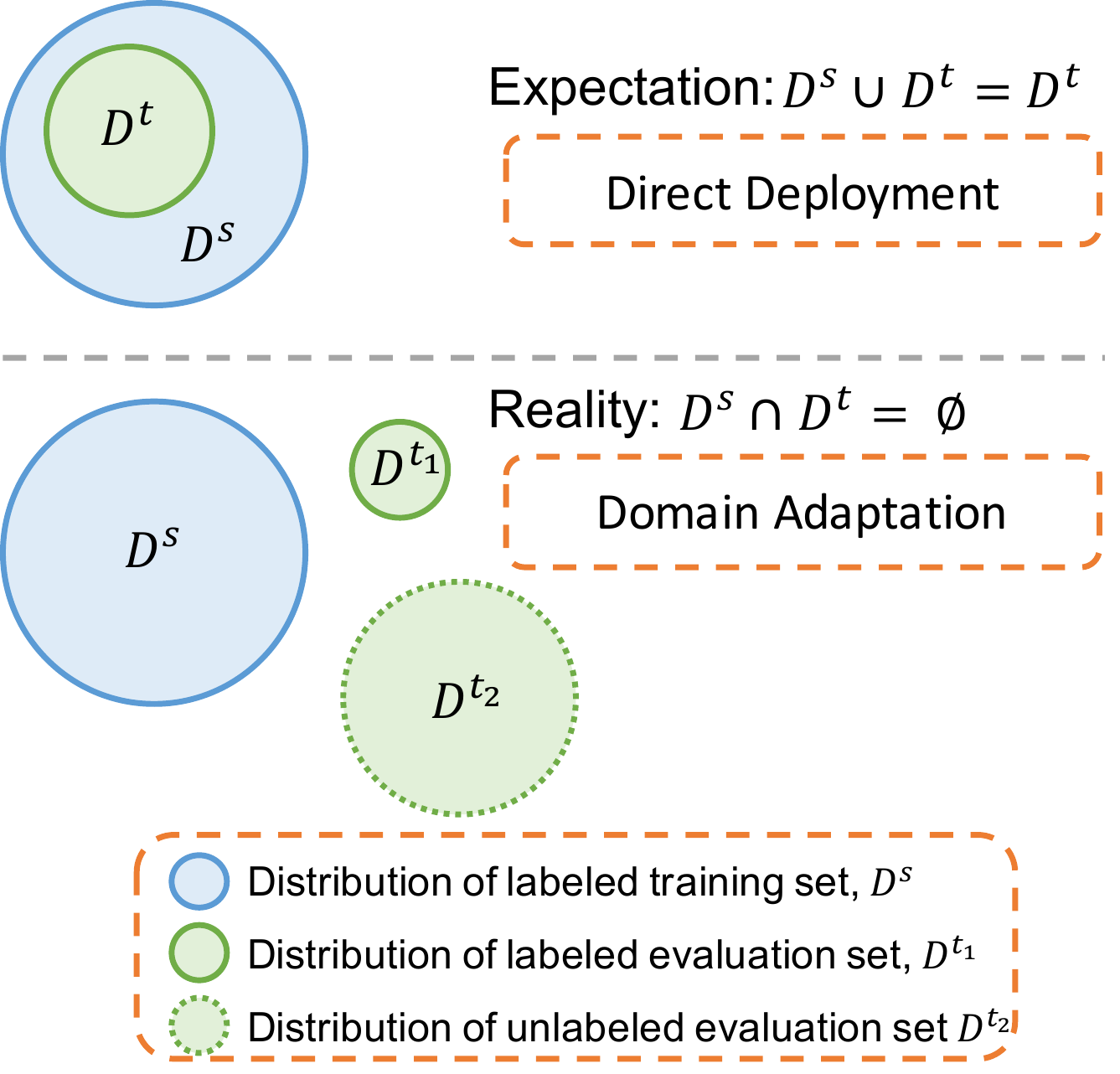}
	\caption{Domain adaptation in the true data space: Expectation vs. Reality.}
	\label{FIG:r-DA-explanation}
\end{figure}

The use of pre-trained models and transfer-learning have become commonplace in today's deep learning-centric computer vision.
Consider a pre-trained model $\model_{\srcd}$ trained using a large dataset $\srcd=\{(\srcx_i, \srcy_i)\}_{i=1}^{\nsrc}$, where $\srcx_i$ is the $i\sth$ sample of the $s\sth$ domain and $\nsrc$ is the number of samples in the $s\sth$ domain. 
Suppose that a typical user has a smaller dataset $\tgtd = \{(\tgtx_j, \tgty_j)\}_{j=1}^{\ntgt}$, with $\ntgt << \nsrc$, on which they want to train their model. 
Also consider that the label spaces are the same, i.e., $\{\srcy, \tgty\} \in [0, 1, \dots c-1]$. 
The user should be able to repurpose the model $\model_{\srcd}$ to work with dataset $\tgtd$. 
Unless the user is extremely lucky as shown in the top case of figure \ref{FIG:r-DA-explanation}, such a deployment will not work. This is due to domain-shift. Features become meaningless and their spaces get transformed, therefore classifier boundaries have to be redrawn.

The class of such problems where the knowledge from another domain is recycled to work to a new target domain is called \emph{domain adaptation}.
If the solution can perform equally-well in both domains, it is called as \emph{domain generalization}. 

Typical choices of dataset for source are large-scale datasets such as ImageNet \cite{Deng2009imagenet}. 
Donahue \etal popularized the idea of repurposing networks trained on this dataset to be used as generic feature extractors \cite{donahue2014decaf}. 
They hypothesized and successfully demonstrated that in many cases, when there is limited labeled data available in the target domain, as long as it contains only \emph{natural images}, the feature extractors learnt from ImageNet are general enough to produce discriminative features.
Follow-up studies have analyzed the transferability of neural networks and the generality of datasets in-detail \cite{yosinski2014transferable, venkatesan2016generality}.
In all these cases, the label-space is considered independently for both domains and the classifier layer of the networks are sanitized.
Domain adaptation improves the performance of an existing model $\model_{\srcd}$  for $\tgtd$ by \emph{adapting} the knowledge of the model learned from $\srcd$ to $\tgtd$ with the assumption that the label spaces are same, and therefore not needing to sanitize the classifier layers \cite{Patel2015visual, Wang2018deep}. There are two different philosophies in which domain adaptation is typically attacked: (i) \textbf{Domain Transformation:} To build a transformation from target data to source domain and reuse the source feature extractor and classifier ($\tgtx \rightarrow \srcx$). 
Consider the GAN-based methods \cite{Bousmalis2017unsupervised, Russo2018from}. 
These work on the input-level and transform samples from the target domain to mimic distributions of source domains.
(ii) \textbf{Latent-Space Transformation:} To build a transformation of features extracted from source and features extracted from target into each other or into a common latent space. 
Since these are working on the conditional feature spaces, these methods are typically supervised.

\begin{figure*}[t]
	\centering
	\includegraphics[width=\linewidth]{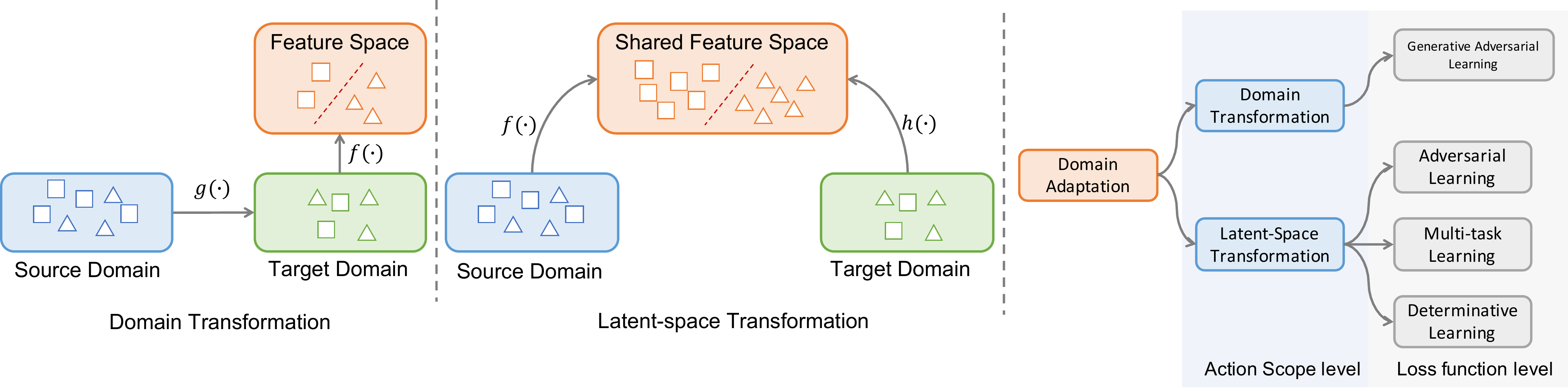}
	\caption{Various types of domain adaptation.}
	\label{FIG:r-DA-types}
\end{figure*} 

Figure \ref{FIG:r-DA-types} illustrates the major branches and types of domain adaptation.
$d$-SNE falls under the latent-space transformation philosophy, where we create a joint-latent embedding space that is agnostic and invariant to domain-shift.
We also focus on the tougher problem where $\ntgt << \nsrc$, or few-shot supervised domain adaptation.
This imposes a constraint that only a few labeled-target samples are available. 

To create this embedding space, we use a strategy that is very similar to the popular stochastic neighborhood embedding technique (SNE) \cite{hinton2003stochastic}.
To modifiy SNE for domain adaptation, we use a novel modified-Hausdorff distance metric in a $\min-\max$ formulation.
$d$-SNE minimizes the distance between the samples from $\srcd$ and $\tgtd$ so as to maximize the margin of inter-class distance for discrimination and minimize the intra-class distance from both domains to achieve domain-invariance.
This discrimination is learnt as a max-margin nearest-neighbor form to make the network optimization easy.
Our proposed idea is still learnable in an end-to-end fashion, therefore making it ideal for training neural networks.

Extensive experimental results in different scenarios indicate that our algorithm is robust and outperforms the state-of-the-art algorithms with only a few labeled data samples.
In several cases, $d$-SNE outperforms even unsupervised methods that have access to all samples in the target domain.
We generalize $d$-SNE such that it can work on a semi-supervised setting that further pushes the states-of-the-art.
Furthermore, $d$-SNE also demonstrates good capabilities in domain generalization without additional training required, which is typically not the case in any state-of-the-art. 

The key contributions in this paper include the following:
\begin{enumerate}
	\item Use of stochastic neighborhood embedding and large-margin nearest neighborhood to learn a domain-agnostic latent-space.
	\item Use of a modified-Hausdorff distance and a novel $\min-\max$ formulation in this space to help few-shot supervised learning.
	\item Demonstration of domain generalization and achieving states-of-the-art results on common benchmark datasets.
	\item Extension to semi-supervised settings pushing the states-of-the-art further.
\end{enumerate}

The rest of this article is organized as follows: section \ref{SEC:related} surveys related literature in all categories described in figure \ref{FIG:r-DA-types}, section \ref{SEC:proposed} dives deep into the proposed idea and provides a theoretical exposition, section \ref{SEC:results} presents validation of this idea through experimental evidence and section \ref{SEC:conclusions} provides concluding remarks.

\section{Related Works}
\label{SEC:related}
In this section we will survey recent related works in each category of domain adaptation.

\noindent \textbf{Domain transformation:} 
In this category, methods learn a generative model that can transform either the source to the target domain (which is more common) or vice-versa.
To learn this generative model itself, no supervision is required.
Since unlabeled data is available in plenty, this model can be learnt easily leveraging a plethora of unsupervised data.
They use this generative model to prepare a joint dataset in either one of the domains and learn a feature extractor and classifier in that \emph{common} domain \cite{Liu2016coupled, Bousmalis2017unsupervised, Hoffman2017cycada, Russo2018from}. 

With the advent of Generative Adversarial Networks (GANs) \cite{goodfellow2014generative} these transformations have become easier
Liu and Tuzel proposed a pair of GANs, each of which was responsible for synthesizing the images in the source and target domain, respectively \cite{Liu2016coupled}.
With an innovative weight-sharing constraint as a regularizer, the generative models were used for generating a pair of images in two domains.
Rather than generating the images from random variables in two domains, the generator in PixelDA, proposed by Bousmalis \etal transformed the images from the source domain and forced them to map into the distribution of the target domain \cite{Bousmalis2017unsupervised}.
CyCADA, proposed by Hoffman \etal used cycle-consistent loss and semantic loss along with the GAN losses and bettered the state-of-the-art \cite{Hoffman2017cycada}.

While all the previous methods transformed the source domain data to target domain, Russo \etal  proposed SBADA-GAN, which also considered the transformation from target to source domain \cite{Russo2018from}.
They defined class consistency loss, which learned to obtain the same label used when mapping from source to target and back tp the source domain.
Since they generated images in both domains, they learnt two independent classifiers in each. 
This implied that they were able to make a prediction using the linear combination or predictions from both domains.

\noindent \textbf{Latent-space transformation:}
Latent-space transformations can be further divided into two major categories: domain adversarial learning (DAL) and domain multi-task learning (DMTL).

\noindent \textit{Domain adversarial learning:} Perhaps the most popular of DAL techniques is the Domain Adversarial Neural Networks (DANN)  introduced by Ganin \etal \cite{Ganin2016domain}.
This work introduced a gradient reversal layer to flip the gradients when the network was back-propagating.
Using this gradient flipping, they were able to learn both a discriminative and a domain-invariant feature space.
The network was optimized to simultaneously minimize the label error and maximize the loss of the domain classifier.
Tzeng \etal generalized the architecture of adversarial domain adaptation for unsupervised learning in their work, Adversarial Discriminative Domain Adaptation (ADDA) \cite{Tzeng2017adversarial}. 
ADDA used two independent discriminators from source and target domain to map features in the shared feature space.
A label-relaxed version of domain adversarial learning was proposed in \cite{cao2018partial}.

\noindent \textit{Domain multi-task learning:}	
In order to improve the discriminative capabilities of feature representations, Tzeng \etal introduced a shared feature extractor for both source and target domain with three different losses in a multi-task learning manner \cite{Tzeng2015simultaneous}.
Ding \etal uses a knowledge graph model to jointly optimize target labels with domain-free features in a unified framework \cite{ding2018graph}.
These losses also acted as a strong regularizer.
Rozantsev \etal argued that the weights of the network learnt from different domain should be related, yet different for each other \cite{Rozantsev2018beyond}.
To this end, they added linear transformations between the weights to regularize the networks to behave thusly.
Associative Domain Adaptation is another technique in the DMTL regime proposed by Haeusser \etal which enforced association between the source and target domains \cite{Haeusser2017associative}.
CCSA and FADA furthered the contrastive loss techniques by creating a unified framework for supervised domain adaptation and generalization \cite{Motiian2017unified, Motiian2017fewshot}.
A decision-boundary iterative refinement training strategy (DIRT-T) was proposed by Shu \etal which required an initialization using virtual adversarial training \cite{Shu2018a}. 
They refined the model's weights with a KL divergence loss.
\emph{Self-ensembling} extended the mean teacher model  in the domain adaptation setting and introduced some tricks such as confidence thresholding, data augmentation, and class imbalance loss \cite{French2018self, Tarvainen2017mean}.
Others learn a shared feature space from the images in the source and target domain \cite{venkateswara2017deep,Sankaranarayanan2018generate} .

%\noindent \textbf{Discussion}: GAN-based methods have the potential to generate a large amount of labeled target similar data. 
%However, they also require a large number of data in the target domain to learn the mapping. 
%In these methods there is also no guarantees that the GAN can generate visually reasonable samples.
%They are much more difficult to train.
%Domain transformation networks on the other hand are generally easier to train.
%Most of recent literature take the unsupervised domain adaptation route, which assumes that the abundance of target unlabeled samples for training.
%This is not an unreasonable assumption to make for \emph{natural images} given the amount of image content available on the internet.
%In specific domains, this will not work.
%Since the adaption of GANs into domain adaptation, supervised and semi-supervised domain adaptation are less studied.
%Our work falls into the DMTL category and focuses on the supervised domain adaptation.

\section{$d$-SNE}
\label{SEC:proposed}

Consider the distance between a sample from the source domain and one from a target domain in the latent-space,
\begin{equation}
\label{eq:distance}
d(x^s_i, x^t_j) = \| \Phi_{\srcd}(x^s_i) - \Phi_{\tgtd}(x^t_j) \|^2_2,
\end{equation}
where $\Phi_{\srcd}(\cdot) \rightarrow \region^d$ and $\Phi_{\tgtd}(\cdot) \rightarrow \region^d$ are neural networks that transform the samples to a common latent-space of $d$-dimensions from the source and target domains, parameterized by $w_s$ and $w_t$ respectively. 
In this latent-space,
\begin{equation}
\label{EQ:SNE}
p_{ij} = \frac{\exp(-d(\srcx_i, \tgtx_j))}{\sum_{x \in \srcd}\exp(-d(x, \tgtx_j))}.
\end{equation}
is the probability that the target sample $\tgtx_j \in \tgtd$ has the same label as the source sample $\srcx_i \in \srcd$.
Since we are working under the supervised regime, we actually have the label for both $\srcx_i$ and $\tgtx_j$, which are $\srcy_i$ and $\tgty_j$, respectively.
If $\srcy_i = \tgty_j$, we want $p_{ij}$ to be maximized.
If otherwise, we want $p_{ij}$ to be minimized.
Notice that in this framework, the training samples in the source domain are chosen from a probability distribution that favors nearby points over faraway ones. 
In other words, the larger the distance between $x$ and $\tgtx_j$, the smaller probability for selecting $x$ as the neighbor of $x^t_j$ for any sample $x \in \srcd$. 

Consider that $\tgty_j= k$ and that $\srcd_k = \{\forall \srcx_l\vert \srcy_l = k\}$.
The probability $p_j$ of making the correct prediction of $\tgtx_j$ is:
\begin{equation}
\begin{aligned}
\label{EQ:tSNE}
p_{j} = \frac{\sum_{x \in \srcd_k}\exp(-d(x, \tgtx_j))}{\sum_{x \in \srcd}\exp(-d(x, \tgtx_j))} = \sum_{i=0}^{\nsrc_k} p_{ij},
\end{aligned}
\end{equation}
where, $\nsrc_k = \vert \srcd_k \vert$.
Notice that given a target sample and label $(x^t_j,y^t_j=k)$, the source domain $\srcd$ is split into two parts as a \emph{same-class set} $\srcd_k$  and a different-class set, $\srcd_{\not k}$.

The denominator in equation~\eqref{EQ:tSNE} can now be decomposed as $\sum_{x \in \srcd_k} \exp(-d(x, \tgtx_j)) + \sum_{x \in \srcd_{\not k}}\exp(-d(x, \tgtx_j))$.  
Given $p_j$ for one sample, the objective function for the domain adaptation problem can be derived as, 
%\begin{equation}
%\label{EQ:dSNE}
%\sum_{x_j \in \tgtd}\frac{1}{p_j} = \sum_{x_j \in \tgtd} \Bigg(\frac{\sum_{x \in \srcd_k}\exp(d(x, x_j))}{\sum_{x\in \srcd_{\not k}}\exp(d(x, x_j))}, \text{ for } k = y_j\Bigg).
%\end{equation}
\begin{equation}
\label{EQ:dSNE}
\sum_{x_j \in \tgtd}\frac{1}{p_j} = \sum_{x_j \in \tgtd} \Bigg(\frac{\sum_{x\in \srcd_{\not k}}\exp(-d(x, x_j))}{\sum_{x \in \srcd_k}\exp(-d(x, x_j))}, \text{ for } k = y_j\Bigg).
\end{equation}

Since we want to maximize the probability $p_j$ of making the correct prediction of $x_j$, we minimize the log-likelihood of $\frac{1}{p_j}$, which is equivalent to minimizing the ratio of intra-class distances to inter-class distances in the latent space. 
\begin{equation}
\label{EQ:likelihood}
\cL=\log \Bigg(\frac{\sum_{x\in \srcd_{\not k}}\exp(-d(x, x_j))}{\sum_{x \in \srcd_k}\exp(-d(x, x_j))}, \text{ for } k = y_j\Bigg).
\end{equation}
%Refer to section \ref{SEC:appendix} for more details on the derivation.

%\subsection*{Relaxation}
\noindent \textbf{Relaxation}:
Since we have sum of exponentials in the likelihood formulation, the ratio in equation \eqref{EQ:likelihood} may have a scaling issue.
This leads to adverse effects in stochastic optimization techniques such as stochastic gradient descent.
Since our feature extractors $\Phi_{\srcd}$ and $\Phi_{\tgtd}$ are neural networks, this is essential.
Therefore, we relax this likelihood with the use of a modified-Hausdorffian distance. 
Instead of optimizing the global distance as in equation \eqref{EQ:likelihood}, we only minimize the largest distance between the samples of the same class and maximize the smallest distance between the samples of different classes.
The final loss is,
\begin{multline}
	\tilde{\cL} = \sup_{x \in \srcd_k} \{ a \vert a \in d(x, x_j) \} - \inf_{x \in \srcd_{\not k}} \{ b \vert b \in d(x,x_j) \},	\\
	\text{ for } k = y_j .
\end{multline}

\begin{figure}[!t]
	\includegraphics[width=\linewidth]{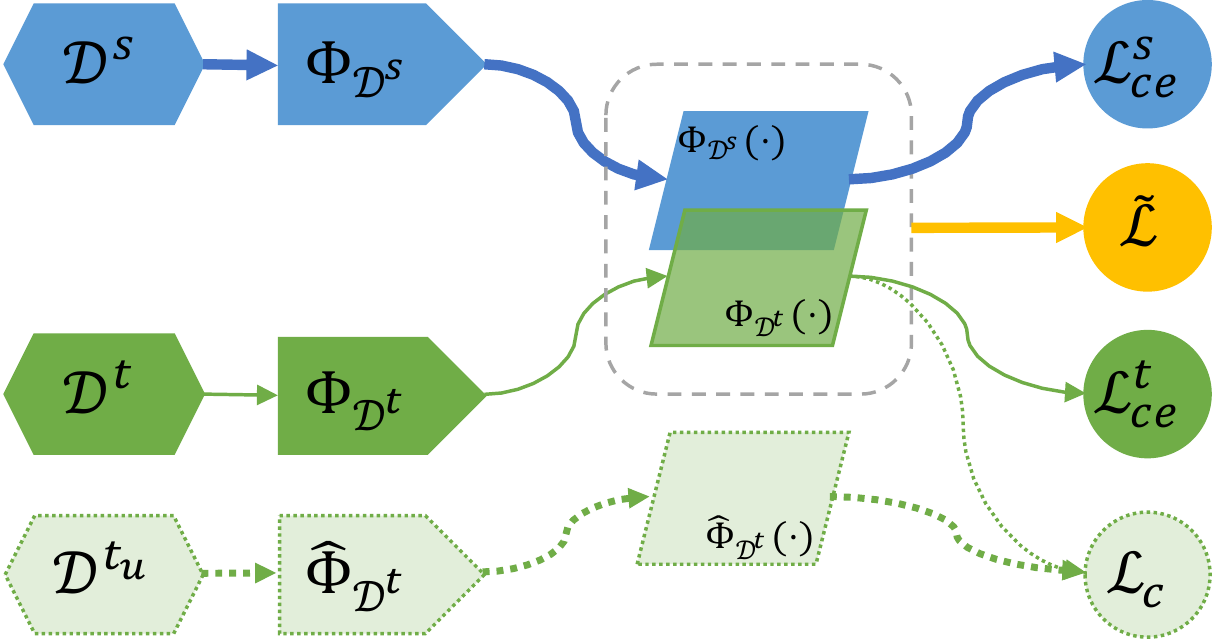}    
	\caption{The learning setup. The segment in the bottom in lighter shade and dotted lines is the semi-supervised extension.}
	\label{FIG:archi}    
\end{figure}

%\subsection*{End-to-End Learning}
\noindent \textbf{End-to-End Learning}:
Our feature extractors are two independent neural networks $\Phi_{\srcd}$ and $\Phi_{\tgtd}$. 
Pragmatically, a single network can be shared between the two domains ($\Phi_{\srcd}=\Phi_{\tgtd}$) if the input data from the source and target domains have the same dimensionality.
$d$-SNE allows the target points to select neighbors from the source domain, therefore, the supervision can be transferred from the source domain to the target domain. 
Since we have labeled data from both domains, standard cross-entropy losses can be used as regularization on top of the domain adaptation losses to train the networks. 
Since each domain gets its own cross-entropy, we create a multi-task setup to learn these networks in parallel.  
Our learning formulation is therefore defined by, 
\begin{equation}
\label{EQ:loss}
\begin{aligned}
\argmin_{w_s, w_d}  \tilde{\cL} + \alpha \cL_{\text{ce}}^s + \beta \cL_{\text{ce}}^t
\end{aligned}
\end{equation}
Although we have one minimization form, we divide them for each network, since the weight updates can be conducted independently.
Figure \ref{FIG:archi} illustrates our setup.

%\subsection*{Semi-supervised Extension}
\noindent \textbf{Semi-supervised Extension}:
As was already discussed in the introduction section, having access to unlabeled data helps boost performance. 
$d$-SNE can be extended easily to accommodate unlabeled data. 
This extends our proposal into a semi-supervised setting.
This is illustrated in the bottom row of figure \ref{FIG:archi}.
Suppose that the unlabeled data from the target domain is represented as $\tgtd_u$. 
We train an unsupervised network $\hat{\Phi}_{\tgtd_u}$, parameterized by $\hat{w_t}$ to produce an embedding for the unsupervised image in the latent space.
Using a technique similar to the Mean-Teacher network technique proposed by Tarvainen et al., \cite{Tarvainen2017mean}.
We use a consistency loss $\cL_c$ across $\hat{\Phi}_{\tgtd_u}$ and $\Phi_{\tgtd}$, by taking an $L_2$ error between the embeddings.

In particular, the source and target networks are first trained as equation~\eqref{EQ:loss}. 
The unlabeled data $\tgtd_u$ from the target domain are then used to train the Mean-Teacher model, where new network networks are initialized with the trained target network $\hat{\Phi}_{\tgtd} \rightarrow \Phi_{\tgtd}$.
To generate inputs for both networks, stochastic augmentations, such as flipping, cropping, color jittering, are used to create two sibling samples.
Since these are two variants of the same sample and belong to the same class, the consistency loss is an error of the embedding.
The weights of $\hat{\Phi}_{\tgtd_u}$ network are updated by back-propagating the consistency loss.
Instead of sharing weights, the weights of $\Phi_{\tgtd}$ are updated with an exponential moving average of the network weights of $\hat{\Phi}_{\tgtd_u}$.

\section{Experiments and Results}
\label{SEC:results}
\begin{figure}[!t]
	\centering
	\includegraphics[width=\linewidth]{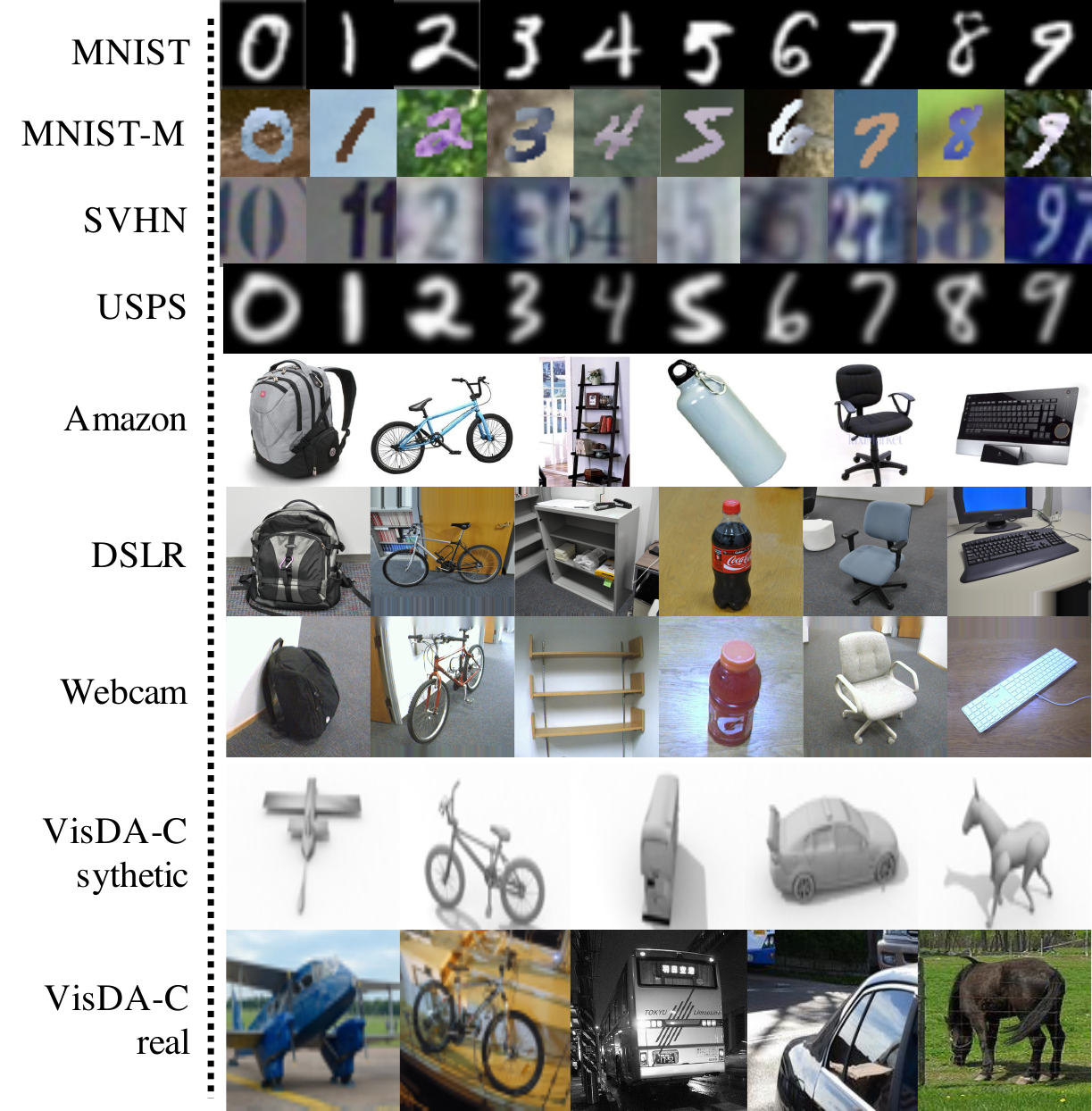}
	\caption{Samples from the datasets used.}
	\label{FIG:data}
\end{figure}

To demonstrate the efficiency of $d$-SNE, three sets of experiments were conducted using three kinds of datasets:
(i) digits datasets \cite{Netzer2011reading}: four datasets are included as different domains in the digits datasets. 
MNIST contains $28 \times 28$ grayscale images with 70,000 images overall.
MNIST-M is a synthetic dataset generated from MNIST by superimposing random backgrounds.
USPS consists of $16 \times 16$ grayscale images, with 9,298 images overall.
SVHN contains RGB photographs of house numbers, with 99,280 images.
(ii) office datasets \cite{Saenko2010adapting}: three sets are included in the office domains. These images are of the same objects but are collected from different sources. Specifically, Office-31 \amazon \ has 2,817 images, which is collected from the Amazon website; 498 images in office-31 \dslr \ are captured by DSLR camera and 795 images in office-31 \webcam \ are captured by web camera.
(iii) VisDA-C dataset \cite{Ganin2016domain}:  two synthetic and real image domains are included in VisDA-C dataset.  152,397 synthetic images are rendered using 3D CAD models as the source domain while the target domain consists of real images.
Figure \ref{FIG:data} shows samples from the datasets used.

\subsection{Digit Datasets}
\begin{table}[!t]
	\centering
	\begin{adjustbox}{width=\linewidth}
		\begin{tabular}{l|cccccc}
			\toprule
			$\vert \tgtd_k \vert , \  \forall k$ &  0 & 1 & 3 & 5 & 7\\
			\midrule \midrule
			CCSA \cite{Motiian2017unified} & $65.40$ & $85.00$ & $90.10$ & $92.40$ & $92.90$\\ 
			FADA \cite{Motiian2017fewshot} & $65.40$ & $89.10$ & $91.90$ & $93.40$ & $94.40$\\
			\midrule
			$d$-SNE &$73.01$ & $\mathbf{92.90}$ & $\mathbf{93.55}$ & $\mathbf{95.13}$ & $\mathbf{96.13}$\\
			\bottomrule
		\end{tabular}    
	\end{adjustbox}
	\caption{MNIST \ $\rightarrow$ USPS \ datasets. $\vert \tgtd_k \vert , \  \forall k$ is essentially number of samples per-class from the target domain. As can be seen, $d$-SNE is clearly able to outperform the states-of-the-art in all scenarios. As the cardinality of the samples per-class increases, the performance across the algorithms converge.}
	\label{TAB:mnist-usps}
\end{table}
\begin{figure*}[!t]
	\includegraphics[width=\linewidth]{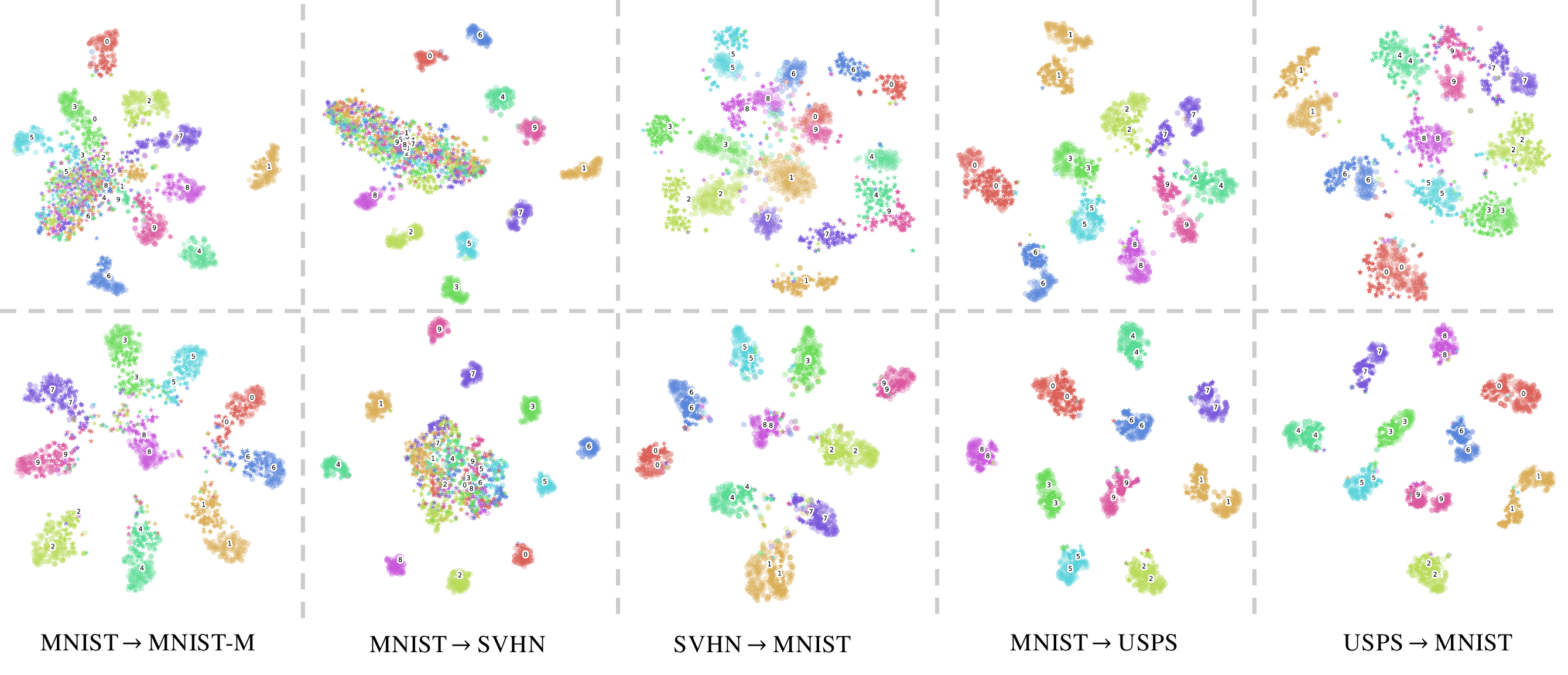}	
	\caption{t-SNE visualizations without (top) and with (bottom) domain adaptations.}
	\label{DA-TSNE-Digits}
\end{figure*}
The first set of experiments adapts the domains of digit datasets.
In the first experiment, the domains considered are MNIST and USPS. 
A total of $2,000$ samples in MNIST are randomly selected for the source domain. 
A small number of samples per-class ranging from $1$ to $7$ were randomly selected from the target domain for training. 
The inputs from the source and target domains have the same dimensionalities, $\Phi_\srcd = \Phi_\tgtd$. 
The states-of-the-art that we use as benchmarks for this experiments are CCSA \cite{Motiian2017unified} and FADA \cite{Motiian2017fewshot}.
We use the same network architecture as them.
Table~\ref{TAB:mnist-usps} shows the overall classification accuracies for adaptation from MNIST to USPS datasets. 
As can be seen that the proposed method outperforms both CCSA and FADA in all the cases even in the one-shot learning case.
For the non-adaptation baseline ($\vert \tgtd_k \vert =0$), it can be noticed that our implementation achieved a higher accuracy than CCSA and FADA. 
We attribute this to a better hyperparameter tuning. 
For the other four cases, we were unable to out-tune their parameters both with our and their own implementations. 
Therefore, we consider their reported numbers as the best for CCSA \cite{Motiian2017unified} and FADA \cite{Motiian2017fewshot}.

\begin{table*}[!t]
	\begin{adjustbox}{width=\linewidth}
		\begin{tabular}{l|c|c|ccccc}
			\toprule
			Method & $\vert \tgtd_k \vert , \  \forall k$ & Setting & MNIST \ $\rightarrow$ MNIST-M & MNIST \ $\rightarrow$ USPS & USPS \ $\rightarrow$ MNIST & MNIST \ $\rightarrow$ SVHN &  SVHN \ $\rightarrow$ MNIST \\
			\midrule \midrule
			PixelDA \cite{Bousmalis2017unsupervised}& &  \multirow{6}{*}{$\mathcal{U}$} & $98.20$ & $95.90$ & - & - & - \\
			ADA \cite{Haeusser2017associative} &    &   &  $87.47$ &  - & - & - & $97.60$ \\
			I2I \cite{murez2018image} &    &   & - &  $95.1$ & $92.2$ & - & $92.1$ \\
			DIRT-T \cite{Shu2018a} & & & $98.90$ & - & - &$54.50$ & $\mathbf{99.40}$\\
			SE \cite{French2018self}& & & - & $98.26 \pm 0.11$ &  $98.07 \pm 2.82$ &$13.96 \pm 4.41$ &$99.18 \pm 0.12$\\
			SBADA-GAN \cite{Russo2018from} & & & $\mathbf{99.40}$ & $95.04$ & $97.60$ & $61.08$  &$76.14$\\
			G2A \cite{Sankaranarayanan2018generate}&  & & -&  $95.30 \pm 0.70$ & $90.80 \pm 1.30$ & - & $92.40 \pm 0.90$\\
			\midrule
			FADA \cite{Motiian2017fewshot} & $7$ & \multirow{2}{*}{$\mathcal{S}$} & - & $94.40$ & $91.50$& $47.00$ & $87.20$ \\
			CCSA \cite{Motiian2017unified} & $10$  & &$78.29 \pm 2.00$ & $97.27 \pm 0.19$ & $95.71 \pm 0.42$ & $37.63 \pm 3.62$ & $94.57 \pm 0.40$ \\
			\midrule
			
			\multirow{3}{*}{$d$-SNE} & $0$ & \multirow{3}{*}{$\mathcal{S}$} & $50.98 \pm 1.64$ & $93.16 \pm 0.71$ & $83.37 \pm 0.93$ & $26.22 \pm 2.02$ & $66.02 \pm 0.72$ \\
			&7 & & $84.62 \pm 0.04$ & $97.53 \pm 0.10$  & $97.52 \pm 0.08$ & $53.19 \pm 0.28$ & $95.68 \pm 0.03$   \\
			&10 & & {$\mathit{87.80 \pm 0.16}$} & $\mathbf{99.00 \pm 0.08}$  & $\mathbf{98.49 \pm 0.35}$ & $\mathbf{61.73 \pm 0.47}$ & $\mathit{96.45 \pm 0.20}$ \\
			\midrule
			$d$-SNE &$10$ & $\mathcal{SS}$ & $94.12$ & - &- & $77.63 \pm 0.26 $&$97.60$  \\						
			\bottomrule
		\end{tabular}
	\end{adjustbox}
	\caption{Classification accuracy for domain adaptation methods on digits datasets. 
		The unsupervised setting ($\mathcal{U}$) uses all the images in the target domain. 
		The supervised setting ($\mathcal{S}$) uses 10 labeled samples per-class from the target domain. 
		We reimplemented CCSA and FADA using the same network and settings as our method.
		The best results are marked in \textbf{bold}. If the best result is not in the supervised-only setting, we mark the best among the supervised-only methods in \textit{italics}.
		The results are averaged over three runs and we report mean and standard deviations over the three runs.}
	\label{DA-EXP-Digits}
\end{table*}

\begin{table}[!ht]
	\begin{adjustbox}{width=\linewidth}
		\begin{tabular}{c|ccc}
			\toprule
			& MNIST $\rightarrow$ MNIST-M & MNIST $\rightarrow$ SVHN & SVHN $\rightarrow$ MNIST \\ 
			\midrule
			before   & 99.45\%                     & 99.51\%                  & 88.96\%                  \\
			after    & 99.51\%                     & 99.59\%                  & 94.94\%                  \\ 
			\bottomrule
		\end{tabular}
	\end{adjustbox}
	\caption{Results of domain generalization.}
	\label{table:DG}
\end{table}
In the second experiment, we used four datasets including MNIST, USPS, MNIST-M, and SVHN  to create five domain adaptation experiments: MNIST $\rightarrow$ MNIST-M, MNIST $\leftrightarrow$ USPS, and MNIST $\leftrightarrow$ SVHN. 
Several states-of-the-art algorithms including, the ones from before use this setup, therefore enabling us to do a lot of comparisons.
It is to be noted that some of the benchmarks are unsupervised, wherein the algorithm uses all of the images in an unlabeled fashion, while the supervised algorithms only use 10 images per-class in the target domain.
We use the same network architecture as Wen et al., \cite{wen2016discriminative}.
The overall classification accuracies are shown in Table~\ref{DA-EXP-Digits}. 
Compared to the other supervised benchmarks, $d$-SNE outperforms the states-of-the-art in all experiments. 
We observed that all supervised methods in general achieved lower accuracies than unsupervised methods in domain pairs MNIST $\rightarrow$ MNIST-M and SVHN $\rightarrow$ MNIST.
In experiments of MNIST$\leftrightarrow$ USPS, $d$-SNE can achieve higher accuracies than even unsupervised methods.
MNIST and USPS datasets has relatively lower intra-class variance compared to MNIST-M and SVHN, which we attribute to these results.
Even though the comparison to unsupervised setting is unfair, we can clearly note that the semi-supervised setting of $d$-SNE pushed our supervised performance closer. 
The methods that outperform us are typically good when using simple datasets.
In Table \ref{TAB:DA-EXP-VISDA17} we can see that with more realistic and complicated datasets, our semi-supervised formulation is also on par and often better than the states-of-the-art.
We suppose this difference in performance to the intuition that digit datasets are easily separable even in unsupervised setting.
Figure \ref{DA-TSNE-Digits} shows some t-SNE visualizations of adaptations.

\noindent \textbf{Domain Generalization}:
%\subsection{Domain Generalization}
As an added benefit, $d$-SNE shows good domain generalization.
Here, we use the model artifacts that we get after the model is adapted to the target domain, and without re-training or fine-tuning, we measure the accuracy on the source dataset.
Table \ref{table:DG} shows classification accuracies on the source domain before and after domain adaptation.
We can notice from Table \ref{table:DG} that the network actually improves the original performance on the source dataset.
This is further evidence of $d$-SNE and the strength of the latent-space it produces.

\begin{table*}[htb]
	\begin{adjustbox}{width=\linewidth}
		\begin{tabular}{l|c|c|ccccccc}
			\toprule
			Method  &$\vert \tgtd_k \vert , \  \forall k$ &Setting  & \texttt{A} $\rightarrow$ \texttt{D}&\texttt{A} $\rightarrow$ \texttt{W} & \texttt{D} $\rightarrow$ \texttt{A} & \texttt{D} $\rightarrow$ \texttt{W} & \texttt{W} $\rightarrow$ \texttt{A }&  \texttt{W} $\rightarrow$ \texttt{D} & Avg.\\
			\midrule \midrule
			DANN \cite{Ganin2016domain} && \multirow{4}{*}{$\mathcal{U}$} & - & $73.00$ & - & $96.40$ & - & $99.20$ & -\\
			DRCN \cite{Ghifary2016deep}  & && $67.10 \pm 0.30$ & $68.70 \pm 0.30$ & $56.00 \pm 0.50$ & $96.40 \pm 0.30$ & $54.09 \pm 0.50$ & $99.00 \pm 0.2$ & 73.60 \\
			kNN-Ad \cite{Sener2016learning} & & & $84.10$ & $81.10$ & $58.30$ & $96.40$ & $63.80$ & $99.20$ & $80.48$\\
			I2I \cite{murez2018image} && & $71.10$ & $75.30$ & $50.10$ & $96.50$ & $52.10$ & $99.60$ & $74.12$\\
			G2A \cite{Sankaranarayanan2018generate} && & $87.70 \pm 0.50$ & $89.50 \pm 0.50$ & $72.80 \pm 0.30$ & $97.90 \pm 0.30$ & $71.40 \pm 0.40$ & $99.8 \pm 0.4$ & $86.50$\\
			\midrule
			SDA \cite{Tzeng2015simultaneous} &$3$& \multirow{3}{*}{$\mathcal{S}$} & $86.10 \pm 1.20$ & $82.70 \pm 0.80$ & $66.20 \pm 0.30$ & $95.70 \pm 0.50$ & $65.00 \pm 0.5$ & $97.60 \pm 0.20$ & $82.22$\\
			FADA \cite{Motiian2017fewshot} &$3$ && $88.20 \pm 1.00$ & $88.10 \pm 1.20$ & $68.10 \pm 0.60$& $96.40 \pm 0.80$ & $71.10 \pm 0.90$ & $97.50 \pm 0.90$ & $84.90$ \\
			CCSA \cite{Motiian2017unified} &$0$& & $61.20  \pm 0.90$ & $62.3 \pm 0.80$ & $58.5 \pm 0.80$ & $80.1 \pm 0.60$ & $51.6 \pm 0.90$ & $95.6 \pm 0.70$ & $68.20$\\
			CCSA \cite{Motiian2017unified} &$3$& & $89.00 \pm 1.20$ & $88.20 \pm 1.00$ & $71.80 \pm 0.50$ & $96.40 \pm 0.80$ & $72.10 \pm 1.00$ & $97.60 \pm 0.40$ & $85.80$\\
			\midrule
			\multirow{2}{*}{$d$-SNE (VGG-16)} &$0$ & \multirow{2}{*}{$\mathcal{S}$} & $62.40 \pm 0.40 $ & $61.49 \pm 0.75$ & $48.92 \pm 1.03$ & $82.24 \pm 1.42$ & $ 47.52 \pm 0.94$ & $90.42 \pm 1.00$ & $65.49$\\
			&$3$    & & $91.44 \pm 0.23$ & $90.13 \pm 0.07$  & $71.06 \pm 0.18$ & $97.10 \pm 0.07$ & $71.74 \pm 0.42$& $97.46 \pm 0.24$& $86.49$\\
			\midrule
			\multirow{2}{*}{$d$-SNE (ResNet-101)} &$0$ & \multirow{2}{*}{$\mathcal{S}$} & $80.41 \pm 0.79$ & $75.26 \pm 1.32$ & $67.39 \pm 0.18$ & $96.39 \pm 0.41$ & $65.55 \pm 1.91$ & $98.31 \pm 1.87$ & $80.55$\\
			&$3$	& & $\mathbf{94.65 \pm 0.38}$ & $\mathbf{96.58 \pm 0.14}$  & $\mathbf{75.51 \pm 0.44}$ & $\mathbf{99.10 \pm 0.24}$ & $\mathbf{74.20 \pm 0.24}$& $\mathbf{100.00 \pm 0.00}$& $\mathbf{90.01}$\\
			\bottomrule
		\end{tabular}
	\end{adjustbox}
	\caption{Results of office31 experiments.  $d$-SNE with ResNet-101 base network achieves the best results with only 3 samples in the target domain while $d$-SNE with VGG-16 base network outperforms the baselines in the majority of cases.}
	\label{TAB:DA-EXP-Office-S}
\end{table*}

\subsection{Office31 Dataset}

For our experiment involving the various domains of the office31 dataset, we followed the same protocol used by Tzeng et al., and Motiian et al., \cite{Tzeng2015simultaneous, Motiian2017fewshot, Motiian2017unified}. 
For the source dataset, we randomly selected 20 samples per-class from \amazon, 8 samples per-class from \dslr \ and \webcam \ domains. 
For the target dataset, 3 samples per-class were selected from all three domains. 
The rest of samples from the target domain were used for evaluation.

We used ResNet-101 as the base network and added two extra dense layers to obtain feature representations \cite{he2016identity}.
The dimensionality of the latent-space was $512$.
As is customary in literature, we used pre-trained weights from ImageNet as initialization \cite{Motiian2017fewshot, Motiian2017unified, Sankaranarayanan2018generate, French2018self}.
Table~\ref{TAB:DA-EXP-Office-S} reports the results of the experiments. 
Even with drastically less data samples, $d$-SNE significantly outperformed all the states-of-the-art benchmarks.
When the domain shift is extremely large, which was the case with in \amazon$\rightarrow$\webcam, \webcam$\rightarrow$\amazon, \amazon$\rightarrow$\dslr \ and \dslr$\rightarrow$\amazon, $d$-SNE shows larger margins compared to baselines.
$d$-SNE can handle domain shift better than other states-of-the-art, especially under limited data conditions.
CCSA and FADA used VGG-16 as their base network.
While ResNets are the preferred base networks contemporaneously, for fair comparisons we also report accuracies of $d$-SNE with VGG-16 as the base network. 
As can be seen from Table \ref{TAB:DA-EXP-Office-S} that the proposed $d$-SNE outperformed both CCSA and FADA in the majority of cases. 
It is worth mentioning that our non-adaptation baseline (($\vert \tgtd_k \vert =0$)) with VGG-16 actually achieved worse results than CCSA and FADA's baselines. 
This highlights the strength of our domain adaptation loss.
\begin{figure*}[!ht]
	\includegraphics[width=0.32\linewidth]{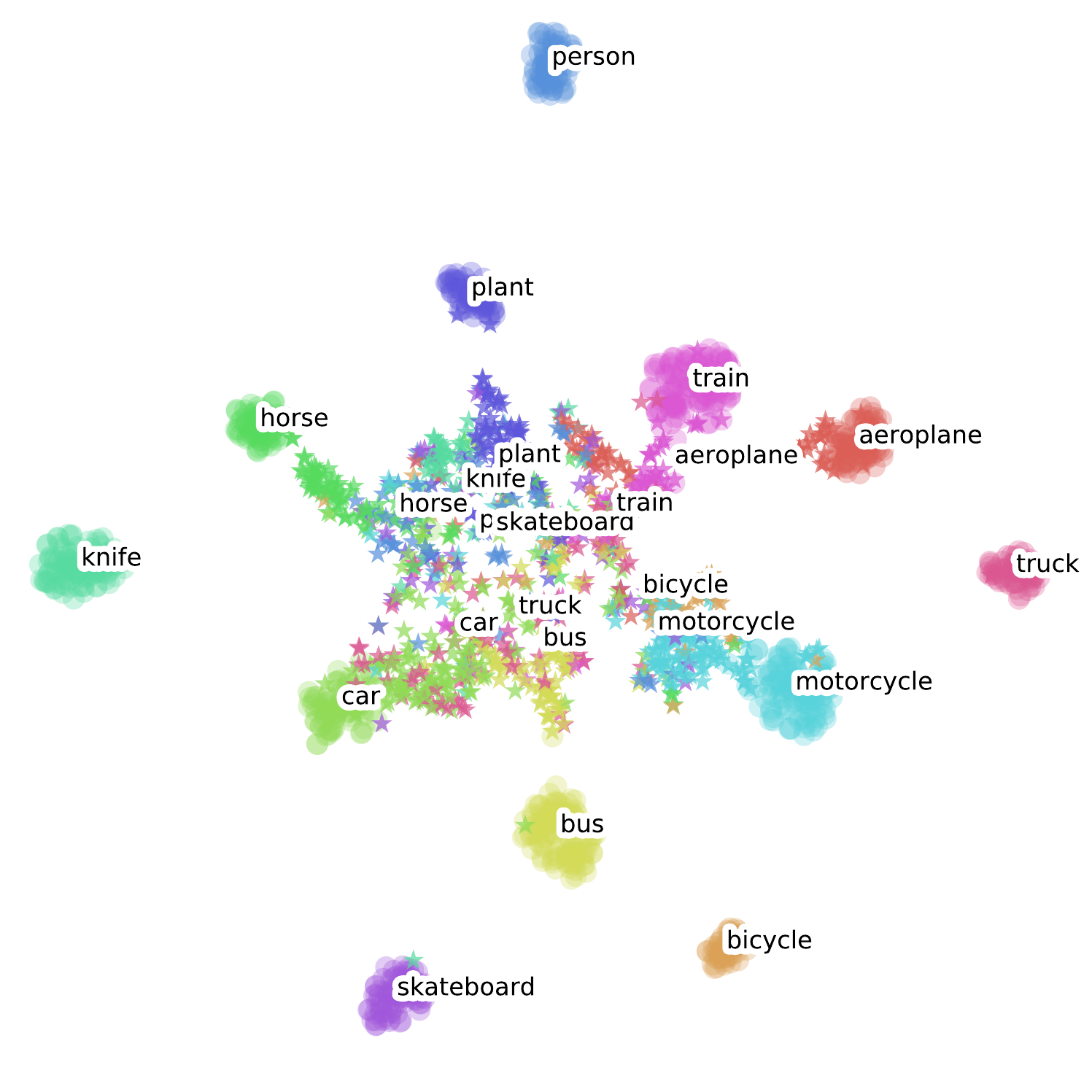}
	\includegraphics[width=0.32\linewidth]{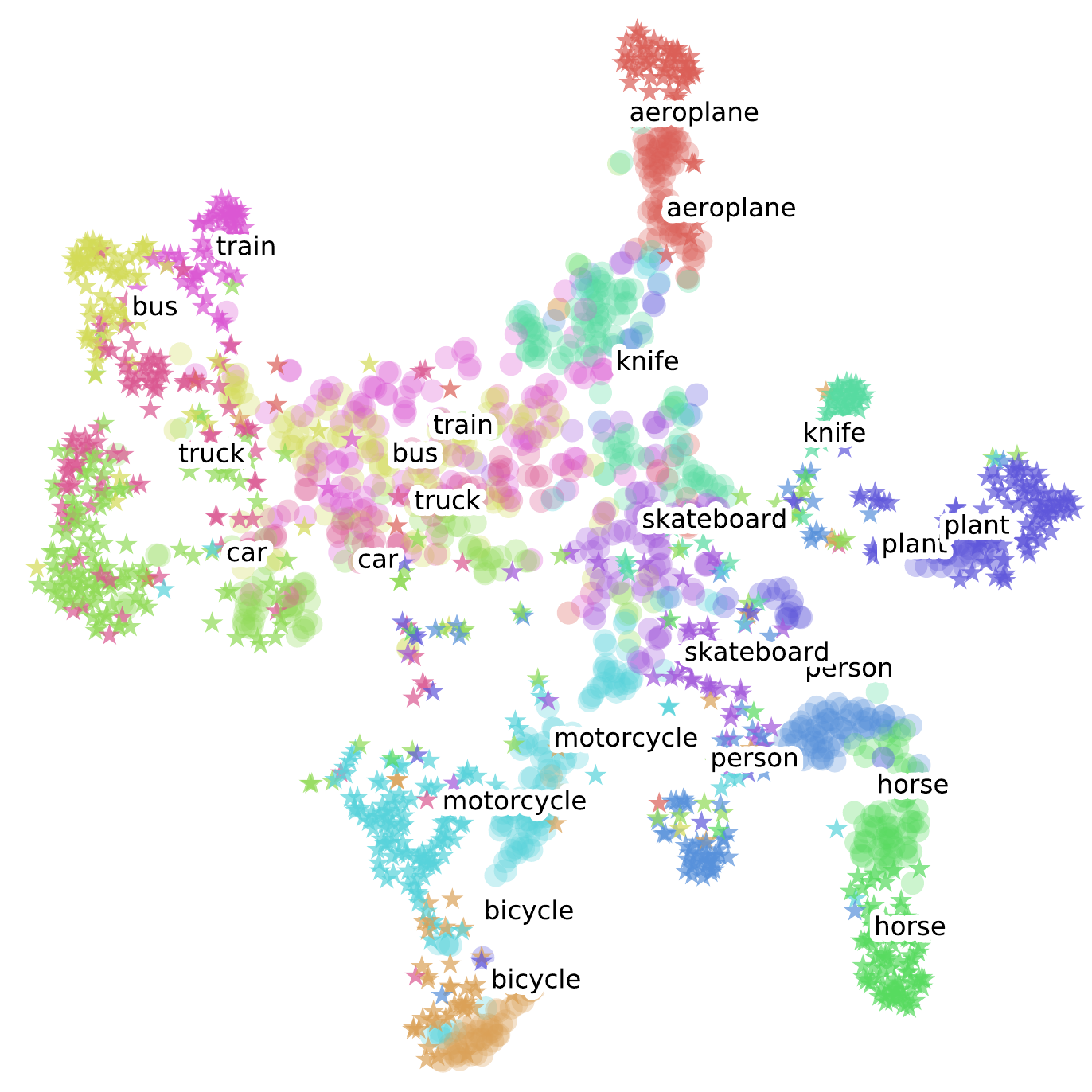}
	\includegraphics[width=0.32\linewidth]{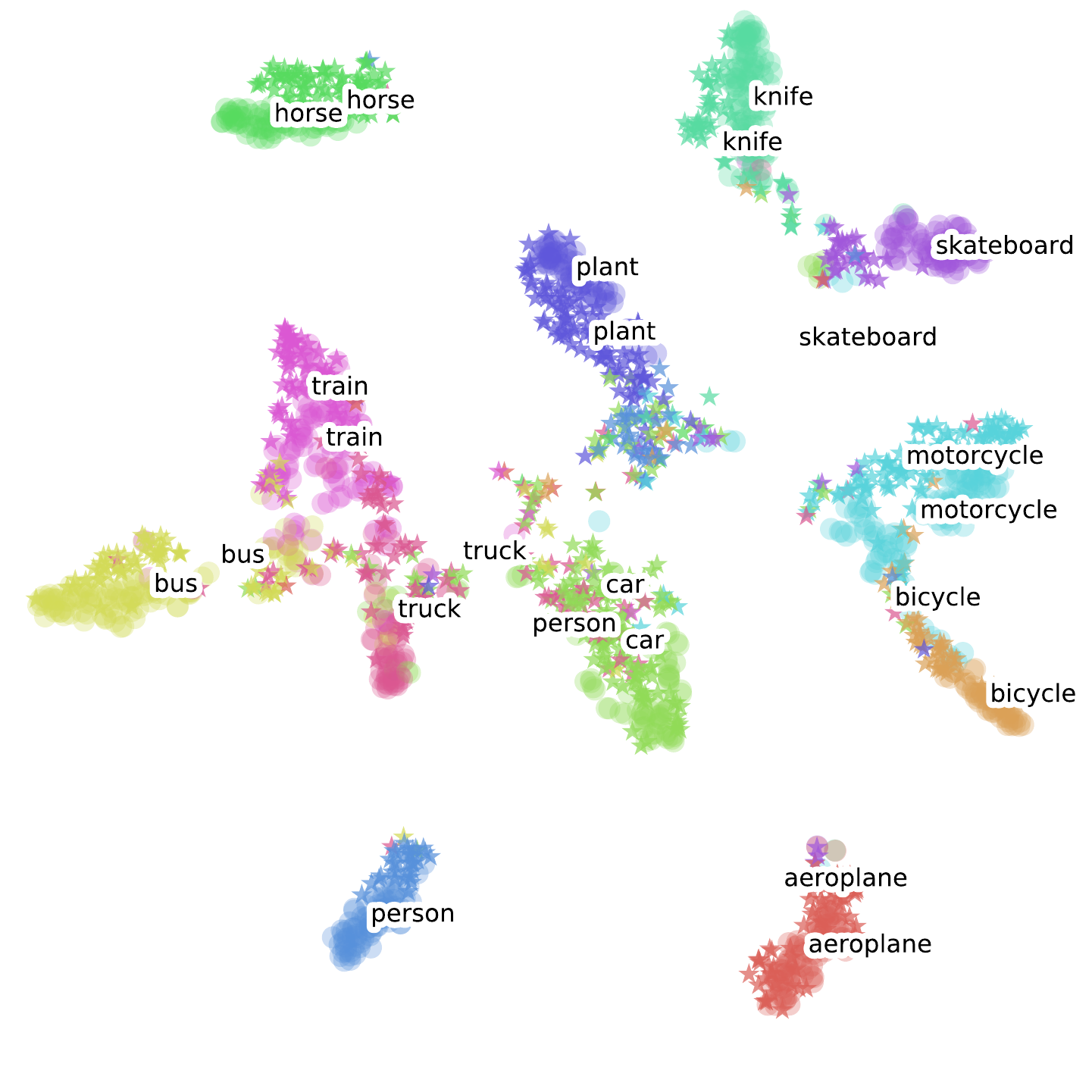}
	\caption{t-SNE visualization of $d$-SNE's latent-embedding space for the VisDA-C dataset. (a) Embeddings produced by the model trained with source images only. (b) Embeddings produced by the model trained with target images only and (c) The joint latent-embedding space of $d$-SNE. Different colors represent different classes. Embeddings from the source and target domains are indicated by circles and stars, respectively.}
	\label{DA-TSNE-VisDA17}
\end{figure*}

\begin{table}[!ht]
	\centering
	\begin{adjustbox}{width=\linewidth}
		\begin{tabular}{lccccc}
			\toprule
			Method & \ \ \ \ &Setting  & $\vert \tgtd_k \vert  = 0, \  \forall k$  & Adaptation\\
			\midrule
			G2A \cite{Sankaranarayanan2018generate} & & $\mathcal{U}$  &$44.50$ & $77.10$ \\
			
			SE \cite{French2018self} & & $\mathcal{SS}$  & $52.80$ & $85.40$ \\
			
			CCSA \cite{Motiian2017unified} & & $\mathcal{S}$ & $52.80$ & $76.89$\\
			\midrule
			\multirow{2}{*}{$d$-SNE}  & & $\mathcal{S}$  & $52.80$ & $80.66$\\
			& & $\mathcal{SS}$  & $52.80$ & $\mathbf{86.15}$\\			
			
			\bottomrule
		\end{tabular}
	\end{adjustbox}
	\caption{Results on the VisDA-C dataset. Source domain was synthetic and target domain was real images with 10 images per-class used for training. 
		The metrics for both G2A and SE were reported from the original source. 
		The results for CCSA were obtained from our own implementation.}
	\label{TAB:DA-EXP-VISDA17}
\end{table}

\subsection{VisDA-C dataset}
VisDA-C is a new dataset, therefore we only have a few reported benchmarks: G2A, SE and CCSA \cite{Sankaranarayanan2018generate, French2018self, Motiian2017unified}. 
G2A is unsupervised, SE is semi-supervised and CCSA is supervised like $d$-SNE. 
With the settings being unique for each algorithm, the results may not be fair, but we include them for comparisons.
We replicated the experimental protocol for the VisDA-C dataset as described in G2A \cite{Sankaranarayanan2018generate}. 

Similar to the baselines, pre-trained models trained on ImageNet were used as initialization.
Table~\ref{TAB:DA-EXP-VISDA17} shows the results on VisDA-C dataset.
With only 10 samples per-class from the target domain for training, $d$-SNE outperformed CCSA and G2A (in unsupervised setting which used all the unlabeled images in the target domain also). 
The supervised $d$-SNE cannot match the performance of the semi-supervised SE, albeit this is not a fair comparison to make. 
With the semi-supervised extension, $d$-SNE established itself as the clear state-of-the-art.
Fig.~\ref{DA-TSNE-VisDA17} demonstrates that the proposed domain adaptation algorithm can align features from both domains while making them discriminative.

\section{Conclusions}
\label{SEC:conclusions}
Domain adaptation has recently seen a massive boom, largely due to the availability of large quantities of data but from varying domains.
Deep-learning-based domain adaptation is a fairly new phenomenon.
In this article, we propose a novel use of the stochastic neighborhood embedding technique-based supervised domain adaptation, that is capable of training neural networks end-to-end.
Experiments across standard benchmarking datasets show that our method establishes a clear state-of-the-art in most datasets.
We propose a semi-supervised extension that pushes the performance further.

{
	\bibliographystyle{ieee}
	\bibliography{DomainAdaptation}
}

\end{document}